\documentclass[10pt,twocolumn,letterpaper]{article}

\usepackage{cvpr}
\usepackage{times}
\usepackage{epsfig}
\usepackage{graphicx}
\usepackage{amsmath}
\usepackage{amssymb}

\usepackage{multirow}

\usepackage[svgnames,table]{xcolor} 
\definecolor{rgb1}{RGB}{214,  38, 40}   
\definecolor{rgb2}{RGB}{43, 160, 4}     
\definecolor{rgb3}{RGB}{158, 216, 229}  
\definecolor{rgb4}{RGB}{114, 158, 206}  
\definecolor{rgb5}{RGB}{204, 204, 91}   
\definecolor{rgb6}{RGB}{255, 186, 119}  
\definecolor{rgb7}{RGB}{147, 102, 188}  
\definecolor{rgb8}{RGB}{30, 119, 181}   
\definecolor{rgb9}{RGB}{160, 188, 33}   
\definecolor{rgb10}{RGB}{255, 127, 12}  
\definecolor{rgb11}{RGB}{196, 175, 214} 

\usepackage[pagebackref=true,breaklinks=true,letterpaper=true,colorlinks,bookmarks=false]{hyperref}

\cvprfinalcopy 

\def\cvprPaperID{4505} 

\ifcvprfinal\fi
\begin{document}

\title{Anisotropic Convolutional Networks for 3D Semantic Scene Completion
\thanks{$ $This work is supported by the National Natural Science Foundation of China under Grants 61773210 and 61603184 and the EPSRC Programme Grant Seebibyte EP/M013774/1.}}



\author{
%
%
Jie~Li$^{1}$  ~~~~Kai~Han$^{2}$  ~~~~Peng~Wang$^{3}$\thanks{~Corresponding author.}  ~~~~Yu~Liu$^{4}$  ~~~~Xia~Yuan$^{1}$
\\
$^{1}$Nanjing University of Science and Technology, China
~~~~$^{2}$University of Oxford, United Kingdom
\\
~~~~$^{3}$University of Wollongong, Australia
~~~~$^{4}$The University of Adelaide, Australia
}

\maketitle

\begin{abstract}
As a voxel-wise labeling task, semantic scene completion (SSC) tries to simultaneously infer the occupancy and semantic labels for a scene from a single depth and/or RGB image. The key challenge for SSC is how to effectively take advantage of the 3D context to model various objects or stuffs with severe variations in shapes, layouts and visibility. To handle such variations, we propose a novel module called anisotropic convolution, which properties with flexibility and power impossible for the competing methods such as standard 3D convolution and some of its variations. In contrast to the standard 3D convolution that is limited to a fixed 3D receptive field, our module is capable of modeling the dimensional anisotropy voxel-wisely. The basic idea is to enable anisotropic 3D receptive field by decomposing a 3D convolution into three consecutive 1D convolutions, and the kernel size for each such 1D convolution is adaptively determined on the fly. By stacking multiple such anisotropic convolution modules, the voxel-wise modeling capability can be further enhanced while maintaining a controllable amount of model parameters. Extensive experiments on two SSC benchmarks, NYU-Depth-v2 and NYUCAD, show the superior performance of the proposed method. Our code is available at
https://waterljwant.github.io/SSC/
.

\end{abstract}

\section{Introduction}

To behave in the 3D physical world, it requires an accurate understanding of both the 3D geometry as well as the semantics of the environment. Humans can easily infer such geometrical and semantic information of a scene from partial observations. An open topic in computer vision is to study how to enable machines such an ability, which is desirable in many applications such as navigation~\cite{doan2019scalable}, grasping~\cite{varley2017shape}, 3D home design~\cite{planner5d}, to name a few.

Semantic scene completion (SSC)~\cite{song2017_SSCNet} is a computer vision task teaching the machine how to perceive the 3D world from the static depth and/or RGB image. The task has two coupled objectives: one is 3D scene completion, which aims at inferring the volumetric occupancy of the scene, and the other is 3D scene labeling, which requires to predict the semantic labels voxel-wisely. As the objects within the physical scene carry severe variations in shapes, layouts, and visibility due to occlusions, the main challenge thereon is how to model the 3D context to learn each voxel effectively.

Recently, promising progress has been achieved for SSC~\cite{song2017_SSCNet,Garbade2018_twoStream,guo2018_VVNet,li2019rgbd,liu2018see} by employing deep convolutional neural networks (CNNs). A direct solution is to use 3D convolutional neural network~\cite{song2017_SSCNet} to model the volumetric context, which consists of a stack of conventional 3D convolutional layers. This solution, however, suffers from apparent limitations. On the one hand, 3D convolution renders a fixed receptive field that does not cater to the variations of the objects. On the other hand, 3D convolution is resource demanding, which causes massive computational and memory consumption. 3D convolution variations~\cite{li2019rgbd,zhang2018efficient} are proposed to address such shortcomings. For example, a lightweight dimensional decomposition network is proposed in~\cite{li2019rgbd} to alleviate the resource consumption, but it still leaves the object variation issue unattended.

In this work, we propose a novel module, termed anisotropic convolution, to model object variation, which properties with flexibility and power impossible for competing methods. In contrast to standard 3D convolution and some of its variations that are limited to the fixed receptive field, the new module adapts to the dimensional anisotropy property voxel-wisely and enables receptive field with varying sizes, a.k.a anisotropic receptive field. The basic idea is to decompose a 3D convolution operation into three consecutive 1D convolutions and equip each such 1d convolution with a mixer of different kernel sizes. The combination weights of such kernels along each 1D convolution are learned voxel-wisely and thus anisotropic 3D context can essentially be modeled by consecutively performing such adaptive 1D convolutions. Although we use multiple kernels, \eg 3, due to the dimensional decomposition scheme, our module is still parameter-economic comparing to the 3D counterpart. By stacking multiple such modules, a more flexible 3D context, as well as an effective mapping function from such context to the voxel output, can be obtained. 

The contributions of this work are as follows:
\vspace{-0.1cm}
\begin{itemize}
\item We present a novel anisotropic convolutional network (AIC-Net) for the task of semantic scene completion. It renders flexibility in modeling the object variations in a 3D scene by automatically choosing proper receptive fields for different voxels. 
\vspace{-0.1cm}
\item We propose a novel module, termed anisotropic convolution (AIC) module, which adapts to the dimensional anisotropy property voxel-wisely and thus implicitly enables 3D kernels with varying sizes. 
\vspace{-0.1cm}
\item The new module is much less computational demanding with higher parameter efficiency comparing to the standard 3D convolution units. It can be used as a plug-and-play module to replace the standard 3D convolution unit.
\vspace{-0.1cm}


\end{itemize}

We thoroughly evaluate our model on two SSC benchmarks. Our method outperforms existing methods by a large margin, establishing the new state-of-the-art. Code will be made available.



\vspace{-0.1cm}
\section{Related Work}
\vspace{-0.1cm}
\subsection{Semantic Scene Completion}
\vspace{-0.1cm}
SSCNet~\cite{song2017_SSCNet} proposed by Song \etal is the first work that tries to simultaneously predict the semantic labels and volumetric occupancy of a scene in an end-to-end network. The expensive cost of 3D CNN, however, limits the depth of the network, which hinders the accuracy achieved by SSCNet. Zhang \etal~\cite{zhang2018efficient} introduced spatial group convolution (SGC) into SSC for accelerating the computation of 3D dense prediction task. However, its accuracy is slightly lower than that of SSCNet.
By combining the 2D CNN and 3D CNN, Guo and Tong~\cite{guo2018_VVNet} proposed the view-volume network (VVNet) to efficiently reduce the computation cost and enhance the network depth.  
Li~\etal~\cite{li2019depth} use both depth and voxels as the inputs of a hybrid network and consider the importance of elements at different positions~\cite{IJCV19Adaptive} while training.

Garbade \etal~\cite{Garbade2018_twoStream} proposed a two-stream approach that jointly leverages the depth and visual information. In specific, it first constructs an incomplete 3D semantic tensor for the inferred 2D semantic information, and then adopts a vanilla 3D CNN to infer the complete 3D semantic tensor. 
Liu \etal~\cite{liu2018see} also used RGB-D image as input and proposed a two-stage framework to sequentially carry out the 2D semantic segmentation and 3D semantic scene completion, which are connected via a 2D-3D re-projection layer. However, their two-stage method can suffer from the error accumulation, producing inferior results. Although significant improvements have been achieved, these methods are limited by the cost of 3D convolution and the fixed receptive fields. Li \etal~\cite{li2019rgbd} introduced a dimensional decomposition residual network (DDRNet) for the 3D SSC task. 
Although it achieves good accuracy with less parameters, it still leaves the limitation of using fixed receptive field unattended.

\subsection{Going Beyond Fixed Receptive Field}
\vspace{-0.1cm}
Most existing models utilize fixed-size kernel to model fixed visual context, which are less robust and flexible when dealing with objects with various sizes.

Inception family~\cite{szegedy2017inception,szegedy2016rethinking,szegedy2015going} take receptive field with multiple sizes into account, and it implements this concept by launching multi-branch CNNs with different convolution kernels. The similar idea appears in atrous spatial pyramid pooling (ASPP)~\cite{chen2017rethinking}, multi-scale information was captured via using several parallel convolutions with different atrous(dilation) rates on the top of feature map. These strategies essentially embrace the idea of multi-scale fusion, and the same fusion strategy is uniformly applied to all the positions. Zhang~\etal~\cite{AAAI20bPixelwise} choose a more suitable receptive field by weighting convolutions with different kernel sizes.

STN~\cite{jaderberg2015spatial} designs a Spatial Transformer module to achieve invariance in terms of translation, rotation, and scale.
However, it treats the whole image as a unit, rather than adjusts the receptive field pixel-wisely. 
Deformable CNN (DCNv1)~\cite{dai2017deformable} attempts to adaptively adjust the spatial distribution of receptive fields according to the scale and shape of the object. Specifically, it utilizes offset to control the spatial sampling. DCNv2~\cite{zhu2019deformable} increases the modeling power by stacking more deformable convolutional layers to improve its modelling ability and proposes to use a teacher network to guide the training process. 
However, DCNv2 still struggles to control the offset in order to focus on relevant pixels only.

Different from the above methods, the proposed AIC module is tailored for 3D tasks, in particular for SSC in this paper. It is capable of handling objects with variations in shapes, layouts and visibility by learning anisotropic receptive field voxel-wisely. At the same time, it achieves trade-off between semantic completion accuracy and computational cost.

\begin{figure*}[t]
\begin{center}
{
\includegraphics[width=0.95\linewidth]{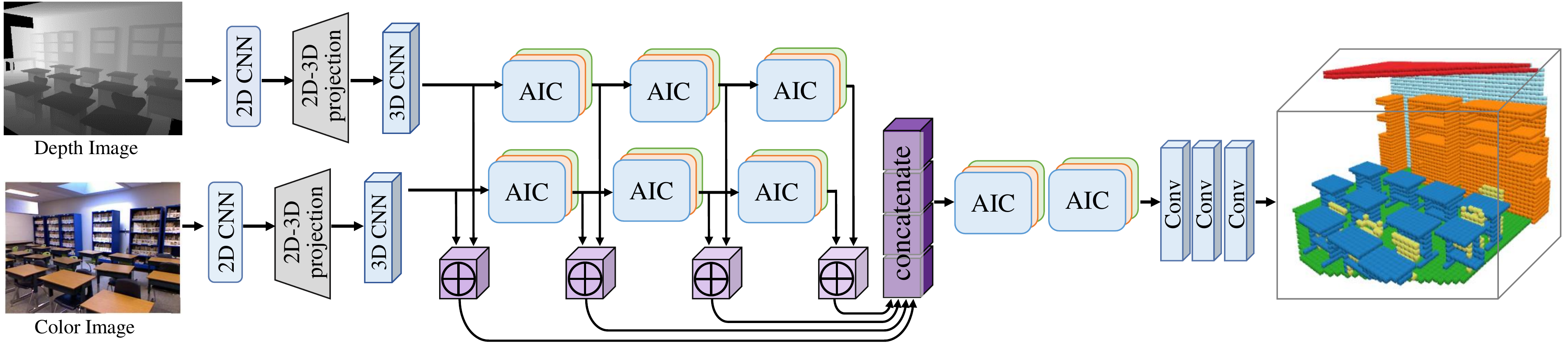}
}
\caption{The overall network structure of AIC-Net. 
AIC-Net has two feature extractors in parallel to capture the features from RGB and depth images, respectively.
The feature extractor contains a projection layer to map the 2D feature to 3D space.
After that, we use stacked AICs to obtain information with adaptive receptive fields.
The multi-scale features are concatenated and then fused through another two AICs followed by three voxel-wise convolutions to predict occupancy and object labels simultaneously.
}
\vspace{-0.6cm}
\label{fig:NetworkStructure}
\end{center}
\end{figure*}
\vspace{-0.1cm}
\section{Anisotropic Convolutional Networks}
\vspace{-0.1cm}
In this section, we introduce our \textit{\textbf{a}n\textbf{i}sotropic  \textbf{c}onvolutional  \textbf{net}works} (AIC-Net) for 3D semantic scene completion. At the core of AIC-Net is our proposed anisotropic convolutional (AIC) module. 
Given a single-view RGB-D image of a 3D scene, AIC-Net 
predicts a dense 3D voxel representation and maps each voxel in the view frustum to one of the labels $C={ \left\{ c_{ 1 }, c_{ 2 },\cdots , c_{ N+1 } \right\}  }$, where $N$ is the number of object classes, $c_{N+1}$ represents the empty voxels, $\{c_1,c_2,\cdots, c_{N}\}$ represent the voxels occupied by objects of different categories.

Fig.~\ref{fig:NetworkStructure} illustrates the overall architecture of our AIC-Net.
It consists of a \emph{hybrid feature extractor} for feature extraction from the depth map and RGB image, a \emph{multi-stage feature aggregation module} with a stack of AIC modules to aggregate features obtained by the hybrid feature extractor, two extra AIC modules to fuse multi-stage information, followed by a sequence of voxel-wise 3D convolution layers to reconstruct the 3D semantic scene.
The \emph{hybrid feature extractor} contains two parallel branches to extract features for the depth map and the RGB image, respectively. Each branch contains a hybrid structure of 2D and 3D CNNs. The 2D and 3D CNNs are bridged by a 2D-3D projection layer, allowing the model to convert the 2D feature maps into 3D feature maps that are suitable for 3D semantic scene completion. The structure of our hybrid feature extractor follows that of DDRNet~\cite{li2019rgbd}. 
The \emph{multi-stage feature aggregation module} consists of a sequence of AIC modules, each of which can voxel-wisely adjust the 3D context on the fly. The outputs of these AIC modules are concatenated together, and another two AIC modules fuse such multi-stage information. 
The 3D semantic scene can then be reconstructed by applying a sequence of voxel-wise 3D convolutional layers on the fused feature.

In the rest of this section, we will introduce our AIC module (section~\ref{sec:aic}), the multi-path kernel selection mechanism achieved by stacking our AIC modules (section~\ref{sec:multipath}), and the training loss for our model (section~\ref{sec:loss}) in detail.



\begin{figure}[t]
\begin{center}   
{
\includegraphics[width=1.0\linewidth]{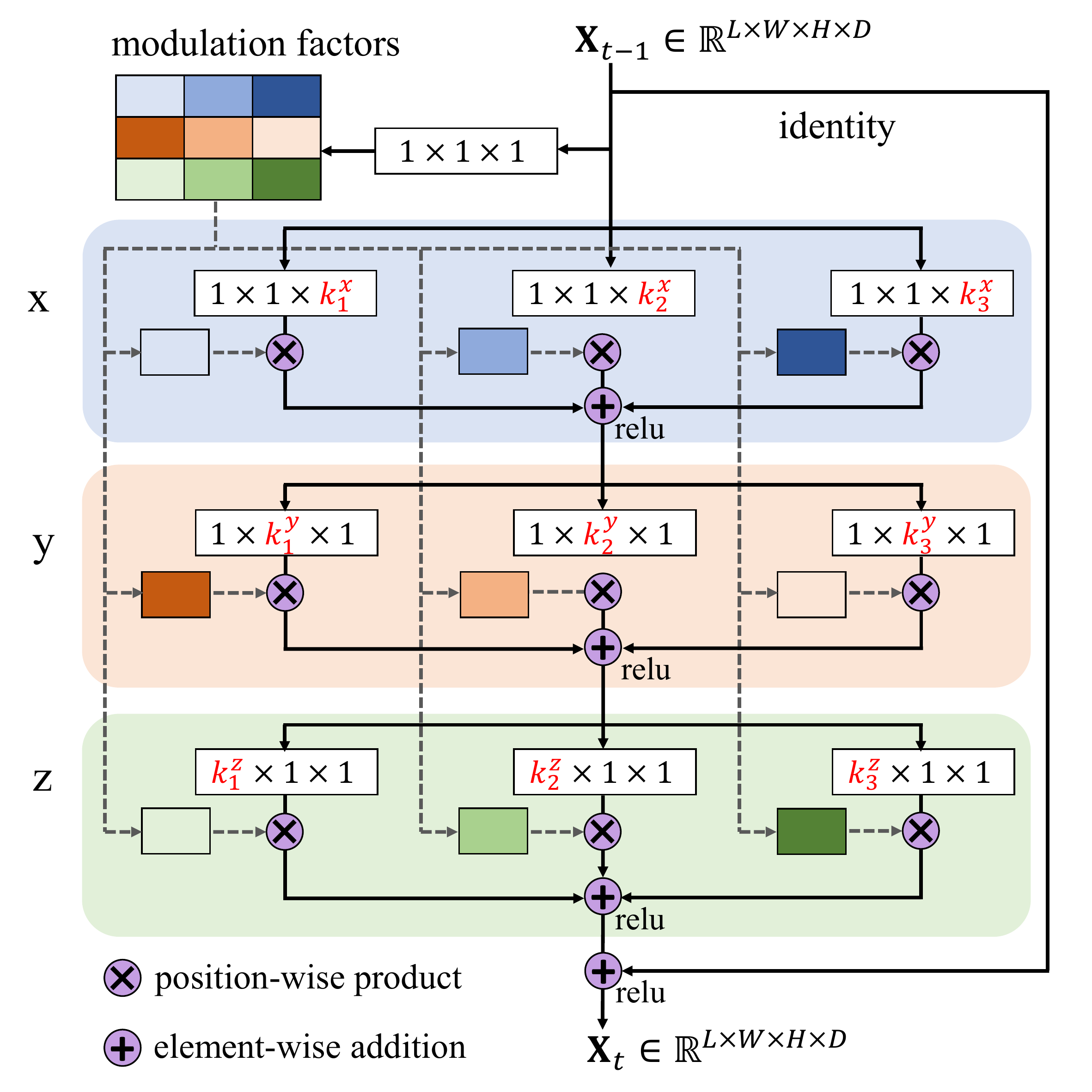}
}
\vspace{-0.2cm}
\caption{Anisotropic convolution. For each dimension, we set 3 parallel convolution with different kernel sizes as an example. The learned modulation factors for different kernels are denoted with different colors. The values of the modulation factors are positive and the values of each row sum up to 1.}
\vspace{-0.4cm}
\label{fig:AIC}
\end{center}
\end{figure}


\subsection{Anisotropic Convolution}
\label{sec:aic}
\vspace{-0.1cm}

Considering the variations in object shapes, layouts as well as the varying levels of occlusion in SSC, it will be beneficial to model different context information to infer the occupancy and semantics for different voxel positions. The anisotropic convolution (AIC) module is proposed to adapt to such variations, allowing the convolution to accommodate 3D geometric deformation.
Fig.~\ref{fig:AIC} shows the structure of our AIC module. Instead of using the 3D kernels ( $k_1\times k_2 \times k_3$) that are limited to the fixed 3D receptive field, we model the dimensional anisotropy property by enabling the kernel size for each 3D dimension to be learnable. To achieve this,
we first decompose the 3D convolution operation as the combination of three 1D convolution operations along each dimension $x$, $y$, $z$. In each dimension, we can inject multiple (\eg 3 in our implementation) kernels of different sizes to enable more flexible context modeling. For example, for dimension $x$, we can have three kernels as $(1 \times 1 \times k_1^x)$, $(1 \times 1 \times k_2^x)$, and $(1 \times 1 \times k_3^x)$. A set of selection weights, a.k.a. modulation factors, will be learned to select proper kernels along each of the three dimensions. Note that the kernel candidates for different dimensions are not necessary to be the same. When there are $n$, $m$, and $l$ candidate kernels along $x$, $y$, and $z$ dimensions respectively, the possible kernel combinations can grow exponentially as,
$\left\{ { k }_{ 1 }^{ z },{ k }_{ 2 }^{ z },\cdots ,{ k }_{ l }^{ z } \right\} \times \left\{ { k }_{ 1 }^{ y },{ k }_{ 2 }^{ y },\cdots ,{ k }_{ m }^{ y } \right\} \times \left\{ { k }_{ 1 }^{ x},{ k }_{ 2 }^{ x },\cdots ,{ k }_{ n }^{ x } \right\}$. 
The AIC module can learn to select different kernels for each dimension, forming an anisotropic convolution to capture anisotropic 3D information.

\begin{figure}[t]
\begin{center}   
{
\includegraphics[width=0.5\linewidth]{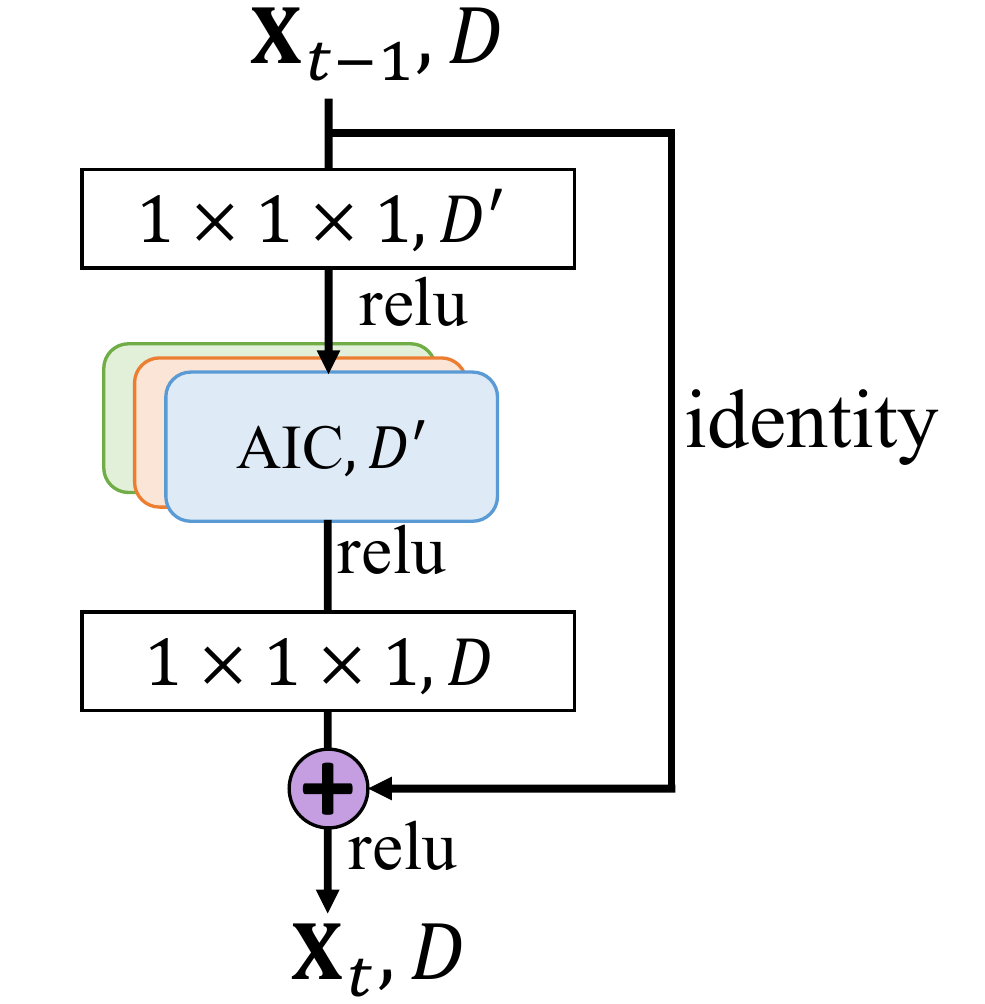}
}
\vspace{-0.1cm}
\caption{Bottleneck version AIC module. The first convolution reduces the number of channels from $D$ to $D'$ ($D' < D$) and the last convolution increases the channels back to $D$.}
\vspace{-0.4cm}
\label{fig:AIC_bottleneck}
\end{center}
\end{figure}


\noindent
\textbf{Modulation factors}
To enable the model to determine the optimal combination of the candidate kernels and consequently adaptively controlling the context to model different voxels, we introduce a modulation module in the AIC module.
As shown in Fig.~\ref{fig:AIC}, assume the input to an AIC module is a tensor $\mathbf{X}_{t-1}\in\mathbb{R}^{L\times{W}\times{H}\times{D}}$, where $L$, $W$, $H$ denotes the length, width, height of the tensor, and $D$ indicates the dimensionality of the feature. The output $\mathbf{X}_{t}\in\mathbb{R}^{L\times{W}\times{H}\times{D}}$ can be formulated as,

\begin{align}
\label{eq:decomp}
\mathbf{X}_t = \mathcal{F}^z(\mathcal{F}^y(\mathcal{F}^x(\mathbf{X}_{t-1})))+\mathbf{X}_{t-1},    
\end{align}
where $\mathcal{F}^u$ represents the anisotropic convolution along the $u\in\{x,y,z\}$ dimension. We adopt a residual structure to obtain the output by element-wisely summing up the input tensor and the output of three consecutive anisotropic 1D convolutions.
Without losing generality, we represent $\mathcal{F}^x(\mathbf{X}_{t-1})$ as,
\begin{align}
\label{eq:kernel_agg}
\mathbf{X}^x_t = \sum^n_{i=1}f^x(\mathbf{X}_{t-1}, \theta^x_i)\odot{g}^x(\mathbf{X}_{t-1}, \phi^{x})[i],
\end{align}
where $f^x(\mathbf{X}_{t-1}, \theta^x_i)$ represents performing convolution to $\mathbf{X}_{t-1}$ using parameter $\theta^{x}_i$ which has kernel size $(1,1,k^x_i)$ with $k^x_i\in\{k^x_1,k^x_2,\cdots,k^x_n\}$, $n$ is the toal numner of candidate kernels for dimension $x$, and $\odot$ denotes element-wise multiplication. $g^x(\mathbf{X}_{t-1}, \phi^x)$ is a mapping function from the input tensor to the weights or modulation factors used to select the kernels along dimension $x$ and $\phi^x$ denotes the parameters of the mapping function. We perform $softmax$ to $g^u(\cdot, \cdot)[i]$ in order that the weights for the kernels of each dimension $u\in\{x,y,z\}$ sum up to 1, that is, 
\begin{align}
    \sum^{p\in\{n,m,l\}}_{i=1}g^{u}(\cdot, \phi^u)[i]=1, \quad g^{u}(\cdot, \phi^u)[i]\geq{0}.
\end{align} 
In this sense, we adopt 
a soft constraint with a set of weights
to determine the importance of different kernels.
The two extreme cases are that the learned modulation factor is $1$ or $0$, indicating that the corresponding kernel will be the unique selected or be ignored. By using soft values, we can control the contributions of these kernels more flexibly.

In Fig.~\ref{fig:AIC}, we show an example of the AIC module with $m=n=l=3$ and as seen,
$g^u(\cdot, \cdot)$ is realized by a 1-layer 3D convolution with kernel $(1\times 1 \times 1)$.

\begin{figure}[t]
\begin{center}
{
\includegraphics[width=1.0\linewidth]{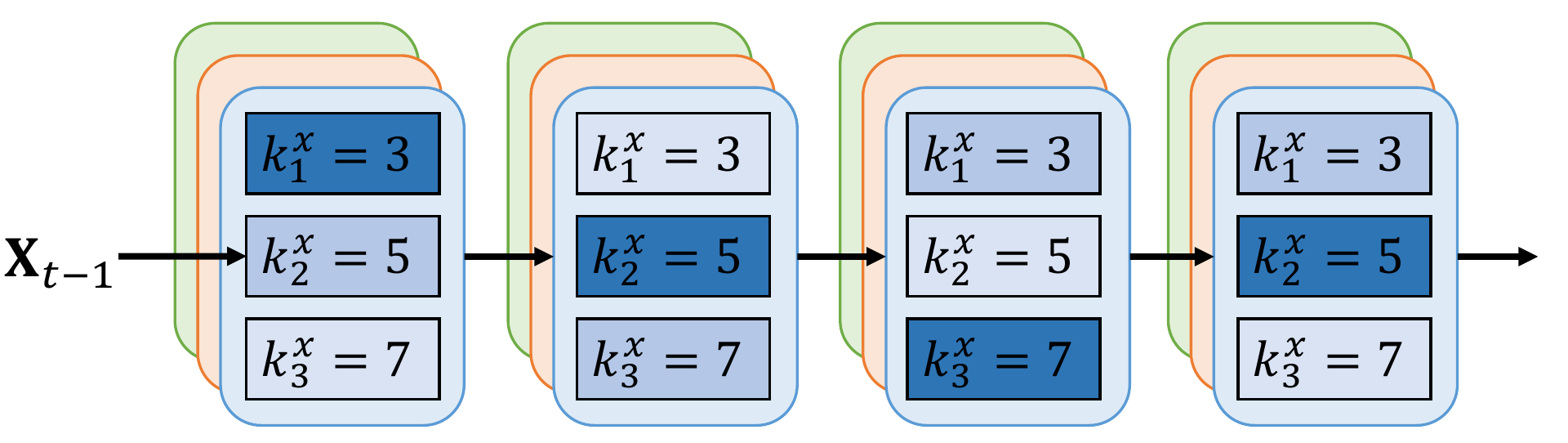}
}
\vspace{-0.2cm}
\caption{Illustration of multi-path kernel selection in one dimension. In this example, four AIC modules are stacked and for each module the kernel sizes for each dimension are $\{3,5,7\}$. The background darkness of the kernel indicates the value of the modulation factor, and thus reflects the selection tendency for this kernel. Stacking multiple AIC modules can increase the range of receptive fields exponentially. 
}
\vspace{-0.4cm}
\label{fig:MultiPath}
\end{center}
\end{figure}

\noindent
\textbf{Bottleneck anisotropic convolution}
To further reduce the parameters of our AIC module, we propose a bottleneck based AIC module. 
As shown in Fig.~\ref{fig:AIC_bottleneck}, for each AIC module, a $(1\times 1\times 1)$ convolution is added both before and after the AIC operation. 
These two convolutions are responsible for reducing and restoring the feature channels, allowing the AIC module to have a more compact input. 
In the remainder of the paper, unless stated otherwise, AIC refers to the bottleneck based AIC.

\begin{table*}[t]
\begin{center}
\scalebox{0.9}
{
\begin{tabular}{l |c c c|c c c c c c c c c c c c} 
\hline
  & \multicolumn{3}{c|}{scene completion} & \multicolumn{12}{c}{semantic scene completion} \\ \hline
Methods  & prec. & recall & IoU & \cellcolor{rgb1}ceil. & \cellcolor{rgb2}floor & \cellcolor{rgb3}wall & \cellcolor{rgb4}win. & \cellcolor{rgb5}chair & \cellcolor{rgb6}bed & \cellcolor{rgb7}sofa & \cellcolor{rgb8}table & \cellcolor{rgb9}tvs & \cellcolor{rgb10}furn. & \cellcolor{rgb11}objs. & avg. \\ 
\hline
Lin \etal~\cite{lin2013holistic}   & 58.5 & 49.9 & 36.4 &  0.0 & 11.7 & 13.3 &  14.1 &  9.4 & 29.0 & 24.0 &  6.0 &  7.0 & 16.2 &  1.1 & 12.0\\
Geiger \etal~\cite{geiger2015joint} & 65.7 & 58.0 & 44.4 & 10.2 & 62.5 & 19.1 &  5.8 &  8.5 & 40.6 & 27.7 &  7.0 &  6.0 & 22.6 &  5.9 & 19.6\\ 
SSCNet~\cite{song2017_SSCNet} & 57.0 & {\bfseries 94.5} & 55.1 & 15.1 & 94.7 & 24.4 &  0.0 & 12.6 & 32.1 & 35.0 & 13.0 &  7.8 & 27.1 & 10.1 & 24.7\\
EsscNet~\cite{zhang2018efficient}   & {\bfseries 71.9} & 71.9 & 56.2 & 17.5 & 75.4 & 25.8 &  6.7 &  15.3 & 53.8 &  42.4 & 11.2 &    0 & 33.4 & 11.8 & 26.7\\ 
DDRNet~\cite{li2019rgbd}  & 71.5  & 80.8 & 61.0 & 21.1 & 92.2 & {\bfseries 33.5} & 6.8 & 14.8 & 48.3 & 42.3 & 13.2 & 13.9 & 35.3 & 13.2 & 30.4\\ 

VVNet~\cite{guo2018_VVNet}  & 69.8 & 83.1 & {\bfseries 61.1} & 19.3 & {\bfseries 94.8} & 28.0 & 12.2 & {\bfseries 19.6} & {\bfseries 57.0} & {\bfseries 50.5} & {\bfseries 17.6} & 11.9 & 35.6 & 15.3 & 32.9
\\ 
\hline
AIC-Net  &62.4&91.8& 59.2& {\bfseries 23.2} & 90.8& 32.3& {\bfseries 14.8} & 18.2 &51.1& 44.8& 15.2&{\bfseries 22.4}&{\bfseries 38.3}&{\bfseries 15.7}&{\bfseries 33.3} 
\\

\hline
\end{tabular}
}

\caption{Results on the NYU~\cite{silberman2012indoor} dataset. Bold numbers represent the best scores.}
\vspace{-0.4cm}
\label{tab:NYU}
\end{center}
\end{table*}

\begin{table*}[t]
\begin{center}
\scalebox{0.9}
{
\begin{tabular} {l |c c c|c c c c c c c c c c c|c} \hline
 &  \multicolumn{3}{c|}{scene completion} & \multicolumn{12}{c}{semantic scene completion} \\ 
\hline
Methods  & prec. & recall & IoU & \cellcolor{rgb1}ceil. & \cellcolor{rgb2}floor & \cellcolor{rgb3}wall & \cellcolor{rgb4}win. & \cellcolor{rgb5}chair & \cellcolor{rgb6}bed & \cellcolor{rgb7}sofa & \cellcolor{rgb8}table & \cellcolor{rgb9}tvs & \cellcolor{rgb10}furn. & \cellcolor{rgb11}objs. & avg. \\ 
\hline
Zheng \etal~\cite{zheng2013beyond} 	& 60.1 & 46.7 & 34.6 & - & - & - & - & - & - & - & - & - & - & - & - \\ 
Firman \etal~\cite{firman2016NYUCAD} 	& 66.5 & 69.7 & 50.8 & - & - & - & - & - & - & - & - & - & - & - & - \\ 
SSCNet~\cite{song2017_SSCNet}   & 75.4 & {\bfseries 96.3} & 73.2 & 32.5 & 92.6 & 40.2 &  8.9 & 33.9 & 57.0 & {\bfseries 59.5} & 28.3 &  8.1 &  44.8 & 25.1 & 40.0\\ 
TS3D~\cite{Garbade2018_twoStream}      	& 80.2 & 91.0 & 74.2 & 33.8 & {\bfseries 92.9} & 46.8 & {\bfseries 27.0} & 27.9 & {\bfseries 61.6} & 51.6 & 27.6 & {\bfseries 26.9} & 44.5 & 22.0 & 42.1\\ 
DDRNet~\cite{li2019rgbd}   & {\bfseries 88.7} & 88.5 & 79.4 & {\bfseries 54.1} & 91.5 & 56.4 & 14.9 & 37.0 & 55.7 & 51.0 &  28.8 & 9.2 & 44.1 &  27.8 & 42.8 \\




VVNet~\cite{guo2018_VVNet}      & 86.4 & 92.0 & 80.3   & - & - & - & - & - & - & - & - & - & - & - & - 
\\ 
\hline
AIC-Net & 88.2 & 90.3 & {\bfseries 80.5} & 53.0  & 91.2 & {\bfseries 57.2} & 20.2 & {\bfseries 44.6} & 58.4 & 56.2 & {\bfseries 36.2}  & 9.7 & {\bfseries 47.1} & {\bfseries 30.4} & {\bfseries 45.8} \\


\hline

\end{tabular}
}  
\caption{Results on the NYUCAD dataset~\cite{zheng2013beyond}. Bold numbers represent the best scores.}
\vspace{-0.7cm}
\label{tab:NYUCAD}
\end{center}
\end{table*}


\subsection{Multi-path Kernel Selection}
\label{sec:multipath}
\vspace{-0.1cm}
Despite the attractive properties in a single AIC module, here we show that greater flexibility can be achieved by stacking multiple AIC modules. Stacking multiple AIC modules forms multiple possible paths between layers implicitly and consequently enables an extensive range of receptive field variations in the model. Fig.~\ref{fig:MultiPath} shows a stack of four AIC modules, and each module sets the kernel sizes to $\{3, 5, 7\}$ along all three dimensions. 
For one specific dimension, when each module tends to select the kernel size 7, a maximum receptive field of $25$ will be obtained for this dimension. On the contrary, a minimum receptive field of $9$ can be obtained for a dimension, if kernel size 3 dominates the selections of all four AIC modules in this dimension. In theory, the receptive field for this particular dimension can freely vary in the range of $(9, 25)$. 
When considering three dimensions simultaneously, the number of 3D receptive fields supported by our AIC network will grow exponentially, which will provide flexibility and power for modeling object variations impossible for competing methods.

\subsection{Training Loss}
\label{sec:loss}
\vspace{-0.1cm}
Our proposed AIC-Net can be trained in an end-to-end fashion. We adopt the voxel-wise cross-entropy loss function~\cite{song2017_SSCNet} for the network training. The loss function can be expressed as,
\begin{equation}\label{Eq:loss}
\mathcal{L} =\sum _{ i,j,k }^{  }{ { w }_{ ijk }{ \mathcal{L} }_{ sm }\left( { p }_{ ijk },{ y }_{ ijk } \right)  },
\end{equation}
where ${ \mathcal{L} }_{ sm }$ is the cross-entropy loss, ${ y }_{ ijk }$ is the ground truth label for coordinates $\left( i, j, k \right) $, ${ p }_{ ijk }$ is the predicted probability for the same voxel, and ${ w }_{ ijk }$ is the weight to balance the semantic categories. We follow~\cite{song2017_SSCNet,li2019rgbd} and use the same weights in our experiments.

\begin{figure*}[t]
\centering
{
\includegraphics[width=0.9\linewidth]{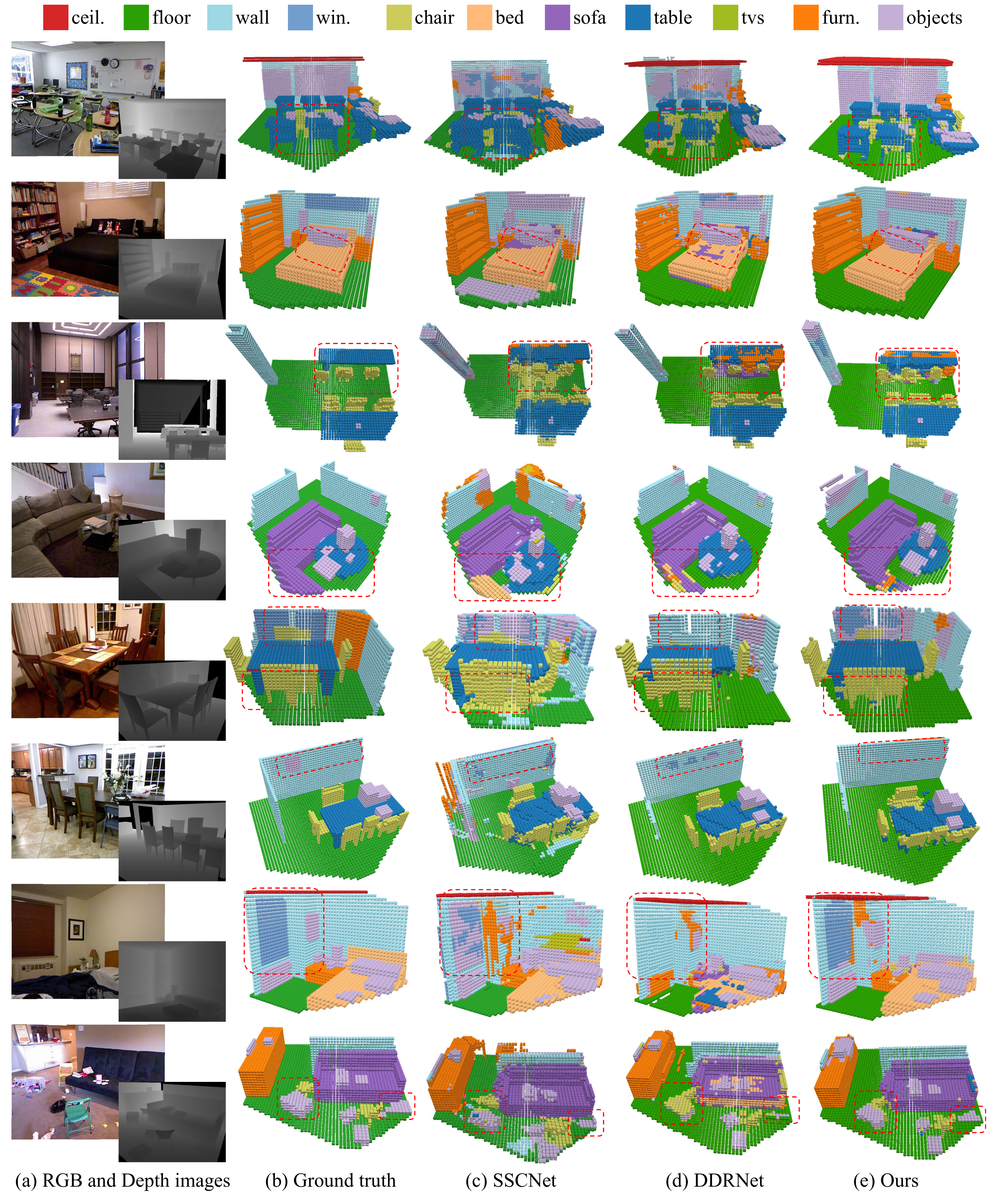}
}
\vspace{-0.1cm}
\caption{Qualitative results on NYUCAD. From left to right are input RGB-D image,
the ground truth, results generated by SSCNet~\cite{song2017_SSCNet}, DDRNet~\cite{li2019rgbd} and the proposed AIC-Net. (Best viewed in color.)
}
\vspace{-0.3cm}
\label{fig:viz_NYUCAD}
\end{figure*}
\section{Experiments}

In this section, we start by introducing some key implementation details, followed by the description of the datasets as well as the evaluation metrics. Then we present some quantitative comparisons between the propose AIC-Net and some other existing works. Furthermore, qualitative comparisons are given through visualization. Finally, comprehensive ablation studies are performed to inspect some critical aspects of AIC-Net.


\subsection{Implementation Details}
\vspace{-0.1cm}
\noindent
In our AIC-Net, we stack three AIC modules for each branch in the multi-stage feature aggregation part, and two AIC modules are adopted to fuse these features. All the AIC modules used are the bottleneck version as shown in Fig.~\ref{fig:AIC_bottleneck}. For the three AIC modules in feature aggregation, the bottleneck layer is used to decrease the dimensionality of the features from $D=64$ to $D'=32$. For the AIC modules in feature fusion part, the dimensionalities of features before and after the bottleneck layer are $D=256$ and $D'=64$.  
Unless stated otherwise, we use three candidate kernels with kernel size $\{3,5,7\}$ for each dimension of all AIC modules. More details about the network structure can be found in the supplements. 

Our model is trained by using SGD with a momentum of 0.9 and a weight decay of $10^{-4}$. The initial learning rate is set to be 0.01, which decays by a factor of 10 every 15 epochs. The batch size is 4. We implement our model using PyTorch. All the experiments are conducted on a PC with 4 NVIDIA RTX2080TI GPUs.

\begin{table*}[t]
\begin{center}
\scalebox{0.88}
{
\begin{tabular} {l |c c c|c c c c c c c c c c c|c} \hline
 &  \multicolumn{3}{c|}{scene completion} & \multicolumn{12}{c}{semantic scene completion} \\ 
\hline
Methods  & prec. & recall & IoU & \cellcolor{rgb1}ceil. & \cellcolor{rgb2}floor & \cellcolor{rgb3}wall & \cellcolor{rgb4}win. & \cellcolor{rgb5}chair & \cellcolor{rgb6}bed & \cellcolor{rgb7}sofa & \cellcolor{rgb8}table & \cellcolor{rgb9}tvs & \cellcolor{rgb10}furn. & \cellcolor{rgb11}objs. & avg. \\ 
\hline
\hline
AIC-Net, $k$=$\{3, 5, 7\}$ & 88.2 & {\bfseries90.3} & {\bfseries 80.5} & {\bfseries53.0}  & 91.2 & {\bfseries 57.2} & 20.2 & {\bfseries 44.6} & 58.4 & 56.2 & {\bfseries 36.2}  & {\bfseries9.7} & {\bfseries 47.1} & 30.4 & {\bfseries 45.8} 
\\

\hline
AIC-Net, $k$=$\{5, 7\}$ & {\bfseries88.3} & 89.5 & 79.9 & 51.0 & 91.3 & 56.8 & 18.6 &  41.3 & {\bfseries58.6} & {\bfseries 59.4} & 34.6  & 4.8 & 46.7 & {\bfseries 30.9} & 44.9  \\
AIC-Net, $k$=$\{7\}$ & 86.3 & {\bfseries 90.3} & 79.1 & 50.7 & {\bfseries 91.7} &54.5 &{\bfseries 21.2} & 38.0 &  55.5& 57.1 &33.2&  7.9& 44.9& 29.4 & 44.0 \\

AIC-Net, $k$=$\{5\}$ & 87.8 & 88.2 & 78.4 & 49.6 & 91.3 & 55.3 & 15.7 & 38.7 & {\bfseries 58.6} & 52.8 & 30.9 &  0. &  43.9 & 30.2 & 42.5 \\
%
\hline

\end{tabular}
}  
\caption{The performance of AIC-Net under different kernel sets. We use the same kernel set $k=(k_1,k_2,\cdots,k_n)$ for each dimension. Results are reported on NYUCAD~\cite{zheng2013beyond} dataset.}
\vspace{-0.3cm}
\label{tab:kernels}
\end{center}
\end{table*}

\begin{table*}[t]
\begin{center}
\scalebox{0.85}
{
\begin{tabular} {l |c c c|c c c c c c c c c c c|c} \hline
 &  \multicolumn{3}{c|}{scene completion} & \multicolumn{12}{c}{semantic scene completion} \\ 
\hline
Methods  & prec. & recall & IoU & \cellcolor{rgb1}ceil. & \cellcolor{rgb2}floor & \cellcolor{rgb3}wall & \cellcolor{rgb4}win. & \cellcolor{rgb5}chair & \cellcolor{rgb6}bed & \cellcolor{rgb7}sofa & \cellcolor{rgb8}table & \cellcolor{rgb9}tvs & \cellcolor{rgb10}furn. & \cellcolor{rgb11}objs. & avg. \\ 
\hline
\hline
NYU  &&&&&&&&&& \\
\hline

AIC-Net-noMFs & {\bfseries71.4} & 79.0 & {\bfseries59.9} & 22.3 & {\bfseries90.8} & 32.0  & 14.4 & 14.5 & 47.5 & 41.3 & 12.6 & 16.8 & 32.8 & 12.7 & 30.7
\\
AIC-Net  &62.4&{\bfseries91.8}&59.2& {\bfseries23.2}& {\bfseries90.8} & {\bfseries32.3} & {\bfseries14.8} & {\bfseries18.2} & {\bfseries51.1} &{\bfseries44.8} &{\bfseries15.2} &{\bfseries22.4} &{\bfseries38.3} &{\bfseries15.7}&{\bfseries33.3} 
\\

\hline
\hline
NYUCAD  &&&&&&&&&& \\
\hline

AIC-Net-noMF & 87.2 & {\bfseries90.3} & 79.6 & 51.1  & {\bfseries91.7}  & 57.0   & 18.5  & 39.3  & 51.4  & 51.8  & 30.7   & 1.3  & 45.0   & 30.1 & 42.5
\\
AIC-Net & {\bfseries88.2} & {\bfseries90.3} & {\bfseries 80.5} & {\bfseries53.0}  & 91.2 & {\bfseries 57.2} & {\bfseries20.2} & {\bfseries 44.6} & {\bfseries58.4} & {\bfseries56.2} & {\bfseries 36.2}  & {\bfseries9.7} & {\bfseries 47.1} & {\bfseries 30.4} & {\bfseries 45.8} 
\\
\hline

\end{tabular}
}  
\caption{The importance of the modulation factors. AIC-Net-noMFs denotes we set all the modulation factors to be 1. Results are reported on the NYU~\cite{silberman2012indoor} and NYUCAD~\cite{zheng2013beyond} datasets.}
\vspace{-0.6cm}
\label{tab:ModulationFactor}
\end{center}
\end{table*}

\noindent
\textbf{Datasets.}
We evaluate the proposed AIC-Net on two SSC datasets. One dataset is the NYU-Depth-V2~\cite{silberman2012indoor}, which is also known as the NYU dataset. The NYU dataset consists of 1,449 depth scenes captured by a Kinect sensor. Following SSCNet~\cite{song2017_SSCNet}, we use the 3D annotations provided by~\cite{rock2015completing} for semantic scene completion task. 
The second dataset is the NYUCAD dataset~\cite{firman2016NYUCAD}. This dataset uses the depth maps generated from the projections of the 3D annotations to reduce the misalignment of depths and the annotations and thus can provide higher-quality depth maps.

\noindent
\textbf{Evaluation metrics.}
For semantic scene completion, we measure the intersection over union (IoU) between the predicted voxel labels and ground-truth labels for all object classes. 
Overall performance is also given by computing the average IoU over all classes. For scene completion, all voxels are to be categorized into either empty or occupied. A voxel is counted as occupied if it belongs to any of the semantic classes. For scene completion, apart from IoU, precision and recall are also reported. Note that the IoU for semantic scene completion is commonly accepted as a more important metric in the SSC task.



\subsection{Comparison with the State-of-the-Art}
\vspace{-0.1cm}
We compare our AIC-Net with the state-of-the-art methods on NYU and NYUCAD. The results are reported in Table~\ref{tab:NYU} and Table~\ref{tab:NYUCAD}, respectively. In Table~\ref{tab:NYU}, we can see that for the semantic scene completion our method significantly outperforms other methods in overall accuracy.
The proposed AIC-Net achieves 2.9\% better than the cutting-edge approach DDRNet~\cite{li2019rgbd} in terms of the average IoU.
For scene completion, our method is slightly outperformed by DDRNet~\cite{li2019rgbd}. The scene completion task requires to predict the volumetric occupancy, which is class-agnostic. Since our AIC-Net aims at modeling the object variation voxel-wisely, its advantage will fade in the binary completion task.
In Table~\ref{tab:NYUCAD}, our AIC-Net achieves the best semantic segmentation performance as well, and our average IoU outperforms the second-best approach by $3\%$. For scene completion, our method also observes superior performance, although the advantage is not as significant. 
Among the comparing methods, SSCNet~\cite{song2017_SSCNet} is built using standard 3D convolution. The inferior performance lies twofold. First, the fixed receptive field is not ideal for addressing object variations. Second, 3D convolution is resource demanding, which can limit the depth of the 3D network and consequently sacrifices the modeling capability.

Another interesting observation from these two tables is that our AIC-Net tends to obtain better performance on some categories that have more severe shape variations, \eg \emph{chair}, \emph{table}, \emph{objects}. 


\subsection{Qualitative Results}
\vspace{-0.1cm}
In Fig.~\ref{fig:viz_NYUCAD}, we show some visualization results to evaluate the effectiveness of our AIC-Net qualitatively.
Generally, we can see that the proposed AIC-Net can handle diverse objects with various shapes and thus give more accurate semantic predictions and shape completion than SSCNet~\cite{song2017_SSCNet} and DDRNet~\cite{li2019rgbd}.   
Some challenging examples include ``chairs'' and ``tables'' in Row 1, Row 3, and Row 5, which require a model to adaptively adjust the receptive field voxel-wisely. For example, for some more delicate parts like ``legs'', a smaller receptive field can be more beneficial. 
It shows that our AIC-Net can identify such objects more clearly.
While for some other objects like ``windows'' in Row 5 and Row 7, it expects to see the larger context. Both SSCNet and DDRNet fail in this case, but our method still successfully identifies them from other surrounding distractors. 
The ``bed'' in Row 2, the ``wall'' in Row 6, and the ``sofa'' in Row 4 also demonstrate the superiority of our approach.
In Row 8, the ``objects'' marked by the red dashed rectangle are in a messy environment. Our AIC-Net is less vulnerable to the influence of surrounding objects and more accurately distinguishes the categories and shapes of these ``objects''.
%

\begin{table*}[t]
\begin{center}
\scalebox{0.88}
{
\begin{tabular} {l |c c c|c c c c c c c c c c c|c} \hline
 &  \multicolumn{3}{c|}{scene completion} & \multicolumn{12}{c}{semantic scene completion} \\ 
\hline
method  & prec. & recall & IoU & \cellcolor{rgb1}ceil. & \cellcolor{rgb2}floor & \cellcolor{rgb3}wall & \cellcolor{rgb4}win. & \cellcolor{rgb5}chair & \cellcolor{rgb6}bed & \cellcolor{rgb7}sofa & \cellcolor{rgb8}table & \cellcolor{rgb9}tvs & \cellcolor{rgb10}furn. & \cellcolor{rgb11}objs. & avg. \\ 
\hline
DDRNet-DDR-ASPP~\cite{li2019rgbd}   &  88.7 & 88.5 & 79.4 & 54.1 & 91.5 & 56.4 & 14.9 & 37.0 & 55.7 & 51.0 &  28.8 & 9.2 & 44.1 &  27.8 & 42.8 \\

DDRNet-AIC-ASPP & 87.9 & 89.1 & 79.4 & 48.0 & 90.9 & 56.1 & 20.1 & 41.6 & 56.6 & 55.0 & 33.1 & 12.6 & 45.3 & 29.0 & 44.4 \\

DDRNet-DDR-AIC &88.0 & 89.6 & 79.7 & 49.0 & 91.4 & 57.6 & 19.7 & 40.5 & 52.3 & 52.9 & 32.5 & 6.1 & 44.6 & 30.7 & 43.4\\

DDRNet-AIC-AIC & 87.5 & 89.3 & 79.1 & 51.7 & 91.5 & 56.4 &   16.5 &   44.1 &  56.3 &  56.4 &  35.4 &  12.3 &  46.1 & 30.4 & 45.2 \\

\hline

\end{tabular}
}  
\caption{AIC module as plug-and-play modules. The components of DDRNet~\cite{li2019rgbd} are replaced by the AIC modules. Results are reported on NYUCAD~\cite{zheng2013beyond} dataset.}
\vspace{-0.6cm}
\label{tab:AIC}
\end{center}
\end{table*}

\begin{table}[t]
\begin{center}
\scalebox{0.8}
{
\begin{tabular} {l|c|c|c|c} 
\hline
Methods 								& Params/k 	& FLOPs/G 	& SC-IoU	& SSC-IoU \\ \hline
SSCNet~\cite{song2017_SSCNet} 			& 930.0  	& 163.8     & 73.2		& 40.0 \\   
DDRNet~\cite{li2019rgbd} 			& 195.0  	& 27.2 	& 79.4	& 42.8 \\ 
\hline




3D conv, $k$=(3, 3, 3) & 440.1 & 61.0 & - & - \\ 

3D conv, $k$=(5, 5, 5) & 1443.6 & 191.1 & - & - \\ 

3D conv, $k$=(7, 7, 7) & 3675.9 & 480.4 & - & - \\ 

\hline

AIC-Net$^{*}$, $k$=$\{3, 5, 7\}$  & 628.7 & 85.5 & 79.1 & 45.2 \\ 

AIC-Net, $k$=$\{3, 5, 7\}$  & 847.0 & 113.7 & 80.5 & 45.8\\ 

AIC-Net, $k$=$\{5, 7\}$  & 716.0 & 96.77 & 79.9 & 44.9 \\

\hline
\end{tabular}
}
\caption{Params, FLOPs and Performance of our approach compared with other methods. 3D conv, $k=(k_1,k_2,k_3)$ denotes we replace our AIC module with a 3D convolution unit with 3D kernel $(k_1,k_2,k_3)$. AIC-Net$^{*}$ denotes a AIC-Net with one AIC module in feature fusion part, while by default we use two AIC modules.}
\vspace{-0.6cm}
\label{tab:params}
\end{center}
\end{table}

\subsection{Ablation Study}
\vspace{-0.1cm}
In this section, we dive into the AIC-Net to investigate its key aspects in detail. Specifically, we try to answer the following questions. 1). Is it beneficial to use multiple candidate kernels along each dimension of the AIC module? 2). Is the performance improvement simply coming from multiple kernels? 3). Will that work if the AIC module is used as a plug-and-play module? 4). The trade-off between SSC performance and cost.


%

\vspace{-0.3cm}
\paragraph{The effectiveness of using multiple kernels}
In our AIC module, we use multiple candidate kernels in each dimension $x,y,z$, and use the learned modulation factors to choose proper kernels along each of these dimensions. 
Since we expect our AIC-Net to be able to deal with objects of varying shapes, the kernels in AIC should be sufficiently distinct.
In our experiments, we set the kernel set to be $\{3,5,7\}$ across all three dimensions. 
The first question needs to be clarified is that will it be enough to use only the maximum kernel, \ie 7 in our network? 
Then, are three kernels better than two? 
From the results of Table~\ref{tab:kernels}, we can see, either two kernels $\{5,7\}$ or three kernels $\{3,5,7\}$ can outperform kernel $7$. Since the maximum receptive field for all these three options is $7$, the results demonstrate the benefits of using multiple kernels. At the same time, three kernels outperform two kernels by about $1\%$ because it renders more flexibility in modeling the context.

\vspace{-0.3cm}
\paragraph{Is it necessary to use modulation factors?}
In the above paragraph, we show the benefit of using multiple kernels along each dimension. However, another question arises that is the improvement simply coming from multiple kernels? In other words, is that necessary to learn modulation factors to adaptively select the kernels voxel-wisely? From Table~\ref{tab:ModulationFactor}, we can see when we discard the modulation factors in AIC modules, the performance of AIC-Net observes obvious degradation on both NYU and NYUCAD datasets. These results show that the superior performance of AIC-Net relies on modeling the dimensional anisotropy property by adaptively selecting proper kernels along each dimension.
To further inspect the anisotropic nature of the learned kernels, we observed the statistical values of the modulation factors and found that: 
1.)~the selected kernel sizes are basically consistent with the object sizes; 
2.)~the modulation values for different voxels vary a lot within one scene; 
3.)~the modulation values among the three separable dimensions have significant variation. This indicates the learned ``3D receptive field'' are anisotropic and adaptive. 

\vspace{-0.3cm}
\paragraph{AIC module used as a plug-and-play module}
Due to its ability to model the anisotropic context, our AIC module is expected to be able to benefit other networks when it is used as a plug-and-play module. To validate this, we choose the DDRNet~\cite{li2019rgbd} as the test-bed, and use the AIC module to replace its building blocks, DDR and ASPP. DDR block models 3D convolution in a lightweight manner with the fixed receptive field. ASPP is a feature fusion scheme commonly used in semantic segmentation to take advantage of the multi-scale context.
Table~\ref{tab:AIC} shows the comparison. 
When we use AIC to replace the DDR module in DDRNet~\cite{li2019rgbd}, the SSC-IoU is improved by $1.6\%$. When we replace ASPP by our AIC module, we still observe a $0.6\%$ improvement in semantic segmentation. Finally, when we replace both DDR and ASPP by AIC, the result can be further boosted. 


\vspace{-0.3cm}
\paragraph{Trade-off in performance and cost}
Since we decompose the 3D convolution into three consecutive 1D convolutions, the model parameters and computation grow linearly with the number of candidate kernels in each dimension. While for standard 3D convolution, the parameters and computation will have cubic growth. Table~\ref{tab:params} presents some comparisons in terms of both efficiency and accuracy. For the \emph{3D conv, $k=(k_1,k_2,k_3)$} in the table, it means we use this particular 3D convolution to replace our AIC module. As can be seen, when the 3D kernel size is $(5,5,5)$, it will result in 3 times of parameters and FLOPs comparing to our AIC-Net. When the kernel size is increased to $(7,7,7)$, the parameter and computation scale will be $8$ times more than ours. DDRNet is a lightweight structure, which consumes the least parameters and has the lowest computation complexity, but it observes a glaring performance gap comparing to our method. Thus, our AIC-Net achieves a better trade-off between performance and cost.

\section{Conclusion}

In this paper, we proposed a novel AIC-Net, to handle the object variations in the semantic scene completion (SSC) task. At the core of AIC-Net is our proposed AIC module, which can learn anisotropic convolutions by adaptively choosing the convolution kernels along all three dimensions voxel-wisely. By stacking multiple such AIC modules, it allows us more flexibly to control the receptive field for each voxel. This AIC module can be freely inserted into existing networks as a plug-and-play module to effectively model the 3D context in a parameter-economic manner. Thorough experiments were conducted on two SSC datasets, and the AIC-Net outperforms existing methods by a large margin, establishing the new state-of-the-art.



\appendix


\section{More Details of AIC-Net}

\subsection{Detailed Architectures}
The details of the proposed network structure are shown in Table~\ref{tab:networkdetails}. 
PWConv represents the point-wise convolution, and it is used to adjust the number of channels of the feature map.
The down-sample layer in our network is composed of a max-pooling layer and a convolution layer with stride set as 2. The outputs of the two layers are concatenated before fed into the subsequent layers.

\subsection{Details of Each AIC Module}
In Table~\ref{tab:networkdetails}, we show the details of the Anisotropic Convolution module (AIC). 
We use three candidate kernels with kernel size $\{3,5,7\}$ for each dimension of all AIC modules.
Since we use bottleneck version AIC, the channel dimension $D'$ within each AIC is lower than the output dimension $D$.
We set $D'=32$ for the first six AIC modules and set $D'=64$ for the last two AIC modules.
The stride and dilation rates of each AIC are all set to 1.

\subsection{2D to 3D Projection}
Each point in depth can be projected to a position in the 3D space. We voxelize this entire 3D space with meshed grids to obtain a 3D volume. In the projection layer, every feature tensor is projected into the 3D volume at the location corresponding to its position in depth. With the feature projection layer, the 2D feature maps extracted by the 2D CNN are converted to a view-independent 3D feature volume.

\begin{table*}[ht]  
\begin{center}
\scalebox{0.95}
{
\begin{tabular}{c|c|c|c|c|c}
\hline
Module                             & Operation        & \multicolumn{1}{l|}{\begin{tabular}[c]{@{}l@{}}Output Size\\ 2D: $Height \times Width\times Channels$\\ 3D: $Length\times Height\times Width\times Channcels$\end{tabular}} & Kernel Size & Stride & Dilation \\ \hline
\multirow{8}{*}{Feature Extractor} & PWConv           & $640\times 480\times 8$                                                                                                                                     & 1           & 1      & 1             \\ \cline{2-6} 
                                   & 2D DDR           & $640\times 480\times 8$                                                                                                                                & 3           & 1      & 1             \\ \cline{2-6} 
                                   & 2D DDR           & $640\times 480\times 8$                                                                                                                                & 3           & 1      & 1             \\ \cline{2-6} 
                                   & 2D-3D Projection & $240\times 144\times 240\times 8$                                                                                                                            & -           & -      & -             \\ \cline{2-6} 
                                   & Down-sample      & $120\times 72\times 120\times 16$                                                                                                                            & 3           & 2      & 1             \\ \cline{2-6} 
                                   & 3D DDR           & $120\times 72\times 120\times 16$                                                                                                                            & 3           & 1      & 1             \\ \cline{2-6} 
                                   & Down-sample      & $60\times 36\times 60\times 64$                                                                                                                            & 3           & 2      & 1             \\ \cline{2-6} 
                                   & 3D DDR           & $60\times 36\times 60\times 64$                                                                                                                              & 3           & 1      & 1             \\ \hline
\multirow{10}{*}{Feature Fusion}    & Add              & $60\times 36\times 60\times 64$                                                                                                                              & 3           & 1      & 1             \\ \cline{2-6} 
                                   & AIC $\times 2$    & $60\times 36\times 60\times 64$                                                                                                                              & \{3,5,7\}   & 1      & 1             \\ \cline{2-6} 
                                   & Add              & $60\times 36\times 60\times 64$                                                                                                                              & 3           & 1      & 1             \\ \cline{2-6} 
                                   & AIC $\times 2$    & $60\times 36\times 60\times 64$                                                                                                                              & \{3,5,7\}   & 1      & 1             \\ \cline{2-6} 
                                   & Add              & $60\times 36\times 60\times 64$                                                                                                                              & 3           & 1      & 1             \\ \cline{2-6} 
                                   & AIC $\times 2$   & $60\times 36\times 60\times 64$                                                                                                                              & \{3,5,7\}   & 1      & 1             \\ \cline{2-6} 
                                   & Add              & $60\times 36\times 60\times 64$                                                                                                                              & 3           & 1      & 1             \\ \cline{2-6} 
                                   & Concatenate      & $60\times 36\times 60\times 256$                                                                                                                             & -           & -      & -             \\ \cline{2-6}
                                   & AIC              & $60\times 36\times 60\times 256$                                                                                                                             & \{3,5,7\}   & 1      & 1             \\ \cline{2-6} 
                                   & AIC              & $60\times 36\times 60\times 256$                                                                                                                             & \{3,5,7\}   & 1      & 1             \\ \hline
\multirow{4}{*}{Reconstruction}    & PWConv           & $60\times 36\times 60\times 128$                                                                                                                             & 1           & 1      & 1             \\ \cline{2-6} 
                                   & PWConv           & $60\times 36\times 60\times 128$                                                                                                                             & 1           & 1      & 1             \\ \cline{2-6} 
                                   & PWConv           & $60\times 36\times 60\times 12$                                                                                                                              & 1           & 1      & 1             \\ \cline{2-6} 
                                   & ArgMax           & $60\times 36\times 60\times 12$                                                                                                                              & -           & -      & -             \\ \hline
\end{tabular}
}
\end{center}
\caption{The details of the proposed (AIC-Net) network architecture. Including module name, layer operation, output size, kernel size, stride and dilation.}
\label{tab:networkdetails}
\end{table*}

\section{More Qualitative Results}
As shown in Fig.~\ref{fig:viz_NYUCAD}, our completed semantic 3D scenes are less cluttered and show a higher voxel-wise accuracy compared to DDRNet[10] and SSCNet[15].

\begin{figure*}[t]
\centering
{
\includegraphics[width=0.95\linewidth]{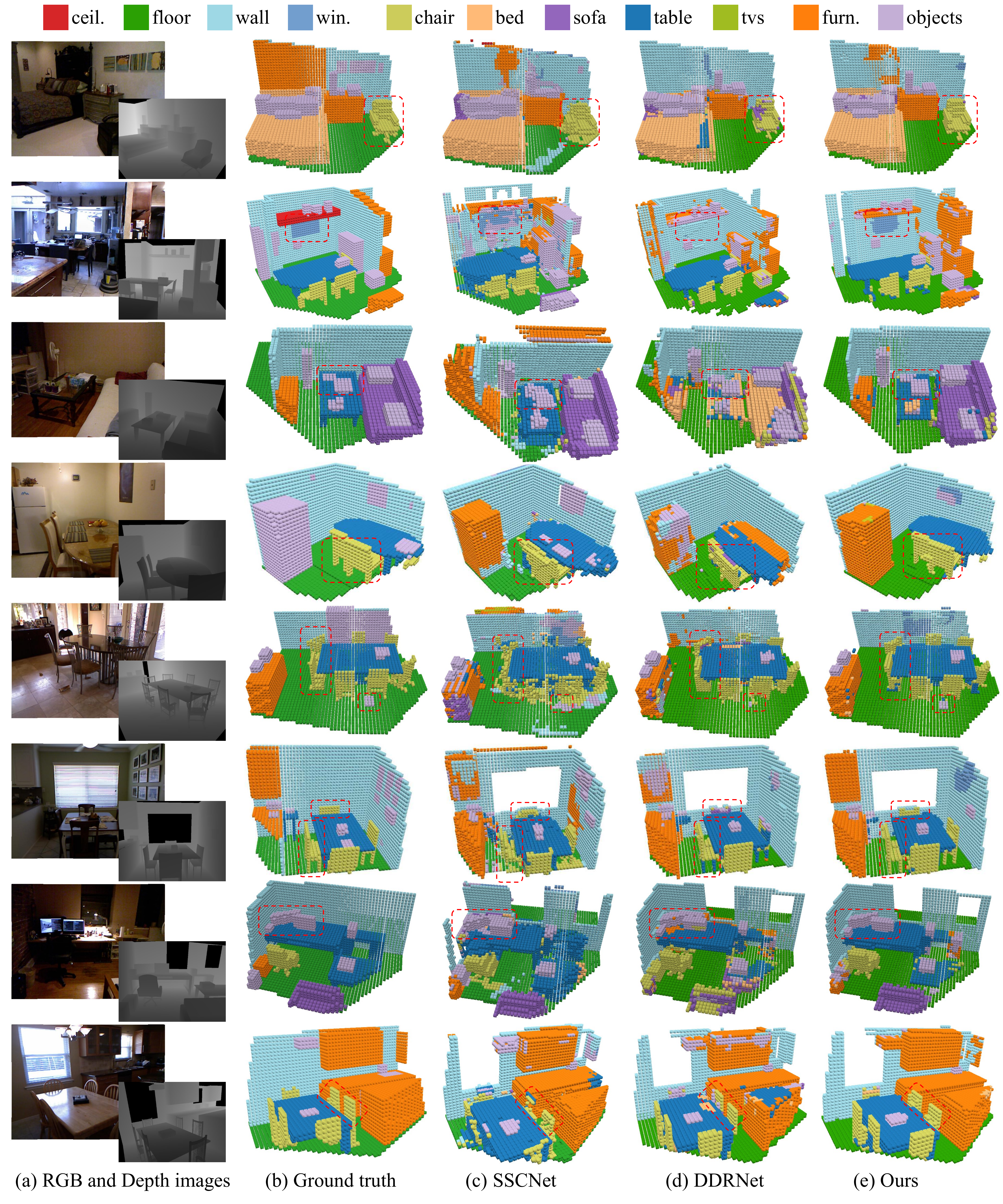}
}
\caption{Qualitative results on NYUCAD[21]. Left to right: input RGB-D image,
the ground truth, results generated by SSCNet[15], DDRNet[10] and the proposed AIC-Net. (Best viewed in color.)
}
\label{fig:viz_NYUCAD_supp}
\end{figure*}

In Fig.~\ref{fig:viz_NYUCAD_supp}, the chair in the first row shows that our result is much more meticulous than the results of the other two methods. In AIC-Net, the irrelevant voxels less interfere with the prediction.
In the second row, the windows are relatively difficult to distinguish, and our method can still distinguish them effectively, while other methods fail.
As shown in rows 3 to 8, the prediction of our AIC-Net is more accurate than other methods. The predicted shape of AIC-Net is more suitable for the actual shape of the object, and the predicted semantic category is more accurate than the other two methods.
We mark the representative areas in Fig.~\ref{fig:viz_NYUCAD_supp} with a red dotted bounding box for easy comparison.

{\small
\bibliographystyle{ieee_fullname}
\bibliography{egbib}

\begin{thebibliography}{10}\itemsep=-1pt

\bibitem{planner5d}
Planner5d.
\newblock \url{https://planner5d.com/}.

\bibitem{chen2017rethinking}
Liang-Chieh Chen, George Papandreou, Florian Schroff, and Hartwig Adam.
\newblock Rethinking atrous convolution for semantic image segmentation.
\newblock {\em arXiv preprint arXiv:1706.05587}, 2017.

\bibitem{dai2017deformable}
Jifeng Dai, Haozhi Qi, Yuwen Xiong, Yi Li, Guodong Zhang, Han Hu, and Yichen
  Wei.
\newblock Deformable convolutional networks.
\newblock In {\em Proceedings of the IEEE international conference on computer
  vision}, pages 764--773, 2017.

\bibitem{doan2019scalable}
Anh-Dzung Doan, Yasir Latif, Tat-Jun Chin, Yu Liu, Thanh-Toan Do, and Ian Reid.
\newblock Scalable place recognition under appearance change for autonomous
  driving.
\newblock In {\em Proceedings of the IEEE International Conference on Computer
  Vision}, pages 9319--9328, 2019.

\bibitem{firman2016NYUCAD}
Michael Firman, Oisin Mac~Aodha, Simon Julier, and Gabriel~J Brostow.
\newblock Structured prediction of unobserved voxels from a single depth image.
\newblock In {\em CVPR}, pages 5431--5440, 2016.

\bibitem{Garbade2018_twoStream}
Martin Garbade, Johann Sawatzky, Alexander Richard, and Juergen Gall.
\newblock Two stream 3d semantic scene completion.
\newblock {\em arXiv:1804.03550}, 2018.

\bibitem{geiger2015joint}
Andreas Geiger and Chaohui Wang.
\newblock Joint 3d object and layout inference from a single rgb-d image.
\newblock In {\em GCPR}, pages 183--195, 2015.

\bibitem{guo2018_VVNet}
Yuxiao Guo and Xin Tong.
\newblock View-volume network for semantic scene completion from a single depth
  image.
\newblock In {\em Proc. IJCAI}, pages 726--732, 7 2018.

\bibitem{jaderberg2015spatial}
Max Jaderberg, Karen Simonyan, Andrew Zisserman, et~al.
\newblock Spatial transformer networks.
\newblock In {\em Advances in neural information processing systems}, pages
  2017--2025, 2015.

\bibitem{li2019rgbd}
Jie Li, Yu Liu, Dong Gong, Qinfeng Shi, Xia Yuan, Chunxia Zhao, and Ian Reid.
\newblock Rgbd based dimensional decomposition residual network for 3d semantic
  scene completion.
\newblock In {\em CVPR}, pages 7693--7702, 2019.

\bibitem{li2019depth}
Jie Li, Yu Liu, Xia Yuan, Chunxia Zhao, Roland Siegwart, Ian Reid, and Cesar
  Cadena.
\newblock Depth based semantic scene completion with position importance aware
  loss.
\newblock {\em IEEE Robotics and Automation Letters}, 5(1):219--226, 2019.

\bibitem{lin2013holistic}
Dahua Lin, Sanja Fidler, and Raquel Urtasun.
\newblock Holistic scene understanding for 3d object detection with rgbd
  cameras.
\newblock In {\em ICCV}, pages 1417--1424, 2013.

\bibitem{liu2018see}
Shice Liu, Yu Hu, Yiming Zeng, Qiankun Tang, Beibei Jin, Yinhe Han, and Xiaowei
  Li.
\newblock See and think: Disentangling semantic scene completion.
\newblock In {\em Advances in Neural Information Processing Systems}, pages
  263--274, 2018.

\bibitem{rock2015completing}
Jason Rock, Tanmay Gupta, Justin Thorsen, JunYoung Gwak, Daeyun Shin, and Derek
  Hoiem.
\newblock Completing 3d object shape from one depth image.
\newblock In {\em CVPR}, pages 2484--2493. IEEE, 2015.

\bibitem{silberman2012indoor}
Nathan Silberman, Derek Hoiem, Pushmeet Kohli, and Rob Fergus.
\newblock Indoor segmentation and support inference from rgbd images.
\newblock In {\em ECCV}, pages 746--760. Springer, 2012.

\bibitem{song2017_SSCNet}
Shuran Song, Fisher Yu, Andy Zeng, Angel~X Chang, Manolis Savva, and Thomas
  Funkhouser.
\newblock Semantic scene completion from a single depth image.
\newblock In {\em CVPR}, pages 190--198, 2017.

\bibitem{szegedy2017inception}
Christian Szegedy, Sergey Ioffe, Vincent Vanhoucke, and Alexander~A Alemi.
\newblock Inception-v4, inception-resnet and the impact of residual connections
  on learning.
\newblock In {\em Thirty-First AAAI Conference on Artificial Intelligence},
  2017.

\bibitem{szegedy2015going}
Christian Szegedy, Wei Liu, Yangqing Jia, Pierre Sermanet, Scott Reed, Dragomir
  Anguelov, Dumitru Erhan, Vincent Vanhoucke, and Andrew Rabinovich.
\newblock Going deeper with convolutions.
\newblock In {\em Proceedings of the IEEE conference on computer vision and
  pattern recognition}, pages 1--9, 2015.

\bibitem{szegedy2016rethinking}
Christian Szegedy, Vincent Vanhoucke, Sergey Ioffe, Jon Shlens, and Zbigniew
  Wojna.
\newblock Rethinking the inception architecture for computer vision.
\newblock In {\em Proceedings of the IEEE conference on computer vision and
  pattern recognition}, pages 2818--2826, 2016.

\bibitem{varley2017shape}
Jacob Varley, Chad DeChant, Adam Richardson, Joaqu{\'\i}n Ruales, and Peter
  Allen.
\newblock Shape completion enabled robotic grasping.
\newblock In {\em Int. Conf. IROS}, pages 2442--2447, 2017.

\bibitem{zhang2018efficient}
Jiahui Zhang, Hao Zhao, Anbang YaoE, Yurong Chen, Li Zhang, and Hongen LiaoE.
\newblock Efficient semantic scene completion network with spatial group
  convolution.
\newblock In {\em ECCV}, pages 733--749, 2018.

\bibitem{AAAI20bPixelwise}
Lei Zhang, Zhiqiang Lang, Peng Wang, Wei Wei, Shengcai Liao, Ling Shao, and
  Yanning Zhang.
\newblock Pixel-wise deep function-mixture network for spectral
  super-resolution.
\newblock In {\em AAAI Conference on Artificial Intelligence}, 2020.

\bibitem{IJCV19Adaptive}
Lei Zhang, Peng Wang, Chunhua Shen, Lingqiao Liu, Wei Wei, Yanning Zhang, and
  Anton van~den Hengel.
\newblock Adaptive importance learning for improving lightweight image
  super-resolution network.
\newblock {\em International Journal of Computer Vision}, pages 1 -- 21, 2019.

\bibitem{zheng2013beyond}
Bo Zheng, Yibiao Zhao, Joey~C Yu, Katsushi Ikeuchi, and Song-Chun Zhu.
\newblock Beyond point clouds: Scene understanding by reasoning geometry and
  physics.
\newblock In {\em CVPR}, pages 3127--3134, 2013.

\bibitem{zhu2019deformable}
Xizhou Zhu, Han Hu, Stephen Lin, and Jifeng Dai.
\newblock Deformable convnets v2: More deformable, better results.
\newblock In {\em Proceedings of the IEEE Conference on Computer Vision and
  Pattern Recognition}, pages 9308--9316, 2019.

\end{thebibliography}
}

\end{document}